\documentclass[10pt,twocolumn,letterpaper]{article}

\usepackage{cvpr}              
\definecolor{cvprblue}{rgb}{0.21,0.49,0.74}
\usepackage[pagebackref,breaklinks,colorlinks,allcolors=cvprblue]{hyperref}

\def\paperID{} 
\def\confName{CVPR}
\def\confYear{2026}

\usepackage{adjustbox}
\usepackage{pifont}
\newcommand{\cmark}{\ding{51}}%
\newcommand{\xmark}{\ding{55}}%
\usepackage{xcolor}
\usepackage{amssymb}
\usepackage{siunitx}
\usepackage{graphicx}
\usepackage{subcaption}
\usepackage{multirow} 
\usepackage{wrapfig}
\usepackage{xspace}
\usepackage{booktabs}

\def\MethodName{FineBench}
\def\MethodNameExt{Benchmarking and Enhancing Vision-Language Models for Fine-grained Human Activity Understanding}

\def\AgentName{FineAgent\xspace}

\title{\MethodName: \MethodNameExt}

\author{
\textbf{Gueter Josmy Faure}$^{1,2}$,  
\textbf{Min-Hung Chen}$^{3}$, 
\textbf{Jia-Fong Yeh}$^{1}$,
\textbf{Hung-Ting Su}$^{1}$,
\textbf{Winston H. Hsu}$^{1}$ \\ \\
$^{1}$National Taiwan University, $^{2}$Google, $^{3}$Independent Researcher \\
}

\begin{document}
\twocolumn[{%
    \renewcommand\twocolumn[1][]{#1}%
    \maketitle
    \begin{center}
        \vspace{-2mm} 
        \begin{minipage}[t]{0.56\textwidth}
            \centering
            \includegraphics[width=\linewidth]{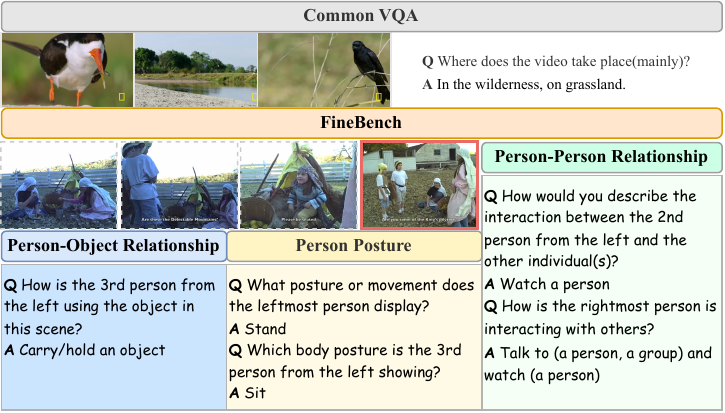}
            \textbf{(a)} \MethodName~versus regular VQAs
        \end{minipage}
        \hfill
        \begin{minipage}[t]{0.42\textwidth}
            \centering
            \includegraphics[width=\linewidth]{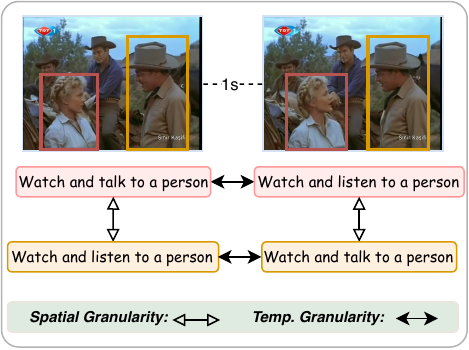}
            \textbf{(b)} \MethodName's Levels of Granularity
        \end{minipage}

        \captionof{figure}{
            \textbf{(a)} Examples of question types in \MethodName{} which go beyond summarization to cover person posture, person-object interaction, and person-person interaction.
            \textbf{(b)} The capture of temporal evolution of interaction labels across frames, emphasizing spatial granularity (e.g., distinguish individuals in the same frame) and temporal granularity (e.g., resolving transitions between similar but distinct actions).
        }
        \label{fig:benchmark_overview}
        \vspace{3mm} 
    \end{center}
}]

\begin{abstract}
    Vision-Language Models (VLMs) have demonstrated remarkable capabilities in general video understanding, yet they often struggle with the fine-grained comprehension crucial for real-world applications requiring nuanced interpretation of human actions and interactions. While some recent human-centric benchmarks evaluate aspects of model behaviour such as fairness/ethics, emotion perception, and broader human-centric metrics, they do not combine long-form videos, very dense QA coverage, and frame-level spatial/temporal grounding at scale. To bridge this gap, we introduce \MethodName, a human-centric video question answering (VQA) benchmark specifically designed to assess fine-grained understanding. \MethodName~comprises 199,420 multiple-choice QA pairs densely annotated across 64 long-form videos (15 minutes each), focusing on detailed person movement, person interaction, and object manipulation, including compositional actions. Our extensive evaluation reveals that while proprietary models like GPT-5 achieve respectable performance, current open-source VLMs significantly underperform, struggling particularly with spatial reasoning in multi-person scenes and distinguishing subtle differences in human movements and interactions. To address these identified weaknesses, we propose \AgentName, a modular framework that enhances VLMs by leveraging a Localizer and a Descriptor. Experiments show that \AgentName~consistently improves the performance of various open VLMs on \MethodName. \MethodName~provides a rigorous testbed for future research into fine-grained human-centric video understanding, while \AgentName~offers a practical approach to enhance such reasoning in current VLMs. 
\end{abstract}    
\section{Introduction}
\label{sec:intro}

Vision-Language Models (VLMs) are rapidly advancing, showing increasing proficiency in interpreting and reasoning about visual content, particularly in the domain of video understanding. However, much of the focus has been on general comprehension tasks—recognizing overall scenes, identifying high-level activities, or summarizing broad events. While valuable, this often falls short in real-world scenarios demanding a \textit{fine-grained} understanding of video content involving humans. Fine-grained video understanding requires perceiving subtle visual details, precise temporal dynamics of actions, complex spatial relationships, and nuanced interactions, especially concerning human behavior. For instance, distinguishing between a person deliberately sitting versus accidentally falling, or discerning intricate social cues in a conversation, requires a level of detail beyond general scene description. Such capabilities are critical for applications ranging from assistive technologies and healthcare monitoring to autonomous systems and detailed behavior analysis.

Despite its importance, fine-grained, human-centric video understanding remains relatively underexplored and under-evaluated in the current VLM landscape. Existing VQA benchmarks often rely on sparsely annotated clips, focus on object-centric or broad activity recognition, or lack the scale and density needed to probe deep, temporally-grounded comprehension \citep{yu2019activitynet, xu2016msr, xiao2021next, li2024mvbench}. As highlighted in Table~\ref{tab:vqa_comparison}, existing benchmarks often lack a specific focus on fine-grained human-centric actions, dense temporal and spatial grounding, or the sheer density of questions required to thoroughly test reasoning over extended video durations. This gap hinders progress, as we lack standardized ways to measure and drive improvements in VLMs' ability to grasp subtle human behavior in videos.

To address this gap, we introduce \textbf{\MethodName{}}, a new benchmark specifically designed to evaluate fine-grained, human-centric video understanding. \MethodName{} is formulated as a multiple-choice VQA dataset containing nearly 200,000 QA pairs derived from 64 long-form videos. Uniquely, it features dense annotations, averaging over 3,100 questions and linking to approximately 785 distinct keyframes per video, enabling detailed assessment of model capabilities at a granular temporal level (e.g., seconds). The questions cover three core domains: Person Movement, Person Interaction, and Object Manipulation, with over 20\% requiring compositional reasoning about combined actions. \MethodName{} explicitly tests spatial and temporal precision through carefully constructed questions and distractors derived from the rich annotations of the AVA v2.2 dataset \citep{gu2018ava}.

Using \MethodName{}, we conduct a comprehensive evaluation of state-of-the-art VLMs, encompassing both leading proprietary models and a wide array of open-source models. Our findings, detailed in Section \ref{subsec:assess} and summarized in Table~\ref{tab:vlm_performance}, reveal a significant performance gap depending on the action type. Rather than a total failure of VLMs, we observe a dramatic performance divide: models excel at \textit{Object Manipulation} tasks (scoring in the high 80s) but perform markedly worse on nuanced \textit{Person Interaction} and \textit{Person Movement} tasks (dropping to the 50s and 60s). While powerful proprietary models achieve a peak accuracy of around 77\%, this indicates there is still over 20\% room for improvement before the benchmark saturates. Furthermore, further analysis (Section \ref{subsec:why}, Figure \ref{fig:radar_analysis}) pinpoints specific weaknesses: VLMs exhibit a marked decline in accuracy as the number of people in the scene increases, underscoring enduring challenges with spatial reasoning and subject disambiguation in challenging multi-person scenes. 

Motivated by these findings, we propose \textbf{\AgentName{}}, a modular framework designed to enhance the fine-grained video understanding capabilities of existing VLMs by directly addressing the identified bottlenecks (Section \ref{sec:fineagent}). \AgentName{} integrates two key components: a \textit{Localizer} that provides explicit bounding box information to aid subject disambiguation in complex scenes, and a \textit{Descriptor} that generates frame summaries, thereby providing richer semantic context. Our main contributions are as follows:
\begin{itemize}
    \item We introduce \MethodName{}, the first densely annotated, human-centric VQA benchmark targeting fine-grained video understanding, featuring 199,420 QA pairs. 
    \item We provide a comprehensive benchmark of current proprietary and open-source VLMs on \MethodName{}, revealing that while models succeed in object-centric tasks, there remains significant room for improvement (over 20\%) in fine-grained reasoning abilities, particularly in spatial reasoning and nuanced action interpretation.
    \item We conduct an in-depth analysis identifying key failure modes for VLMs: degraded performance in multi-person scenarios (spatial reasoning) and difficulties understanding nuanced human movements and interactions.
    \item We propose \AgentName{}, a modular framework leveraging spatial grounding and contextual captioning, demonstrating its effectiveness in improving the fine-grained video understanding performance of various open-source VLMs by targeting their specific weaknesses. 
\end{itemize}

\begin{table*}[htbp]
\centering
\caption{Comparison of \MethodName~with existing VQA datasets across key dimensions. Our dataset is the first to combine fine-grained actions, dense temporal grounding (Temporal G.), dense spatial grounding (Spatial G.), and large-scale QA in a human-centric setting.}
\label{tab:vqa_comparison}
\begin{adjustbox}{max width=\textwidth}
\begin{tabular}{lcccccccccc}
\toprule
 & \textsl{Num. QAs} & \textsl{Num. Videos} & \textsl{Avg. Duration (s)} & \textsl{Density} & \textsl{Human-Centric} & \textsl{Spatial G.} & \textsl{Temporal G.} \\
\midrule
VideoMME \citep{fu2024video}         & 2,700   & 900 & 1017.9 & 3 & \textcolor{red}{\xmark} & \textcolor{red}{\xmark} & \textcolor{red}{\xmark} \\
EgoSchema \citep{mangalam2023egoschema}        & 5,063   & 5,063 & --   & 1 & \textcolor{green!60!black}{\cmark} & \textcolor{red}{\xmark} & \textcolor{red}{\xmark} \\
MovieChat-1k \citep{song2024moviechat}    & 13,000  & 1000 & 564  & 13  & \textcolor{red}{\xmark} & \textcolor{red}{\xmark} & \textcolor{red}{\xmark} \\
ActivityNet-QA \citep{yu2019activitynet}   & 8,000   & 800  & 111.4 & 10 & \textcolor{red}{\xmark} & \textcolor{red}{\xmark} & \textcolor{orange!80!black}{Partial} \\
LongVideoBench \citep{wu2024longvideobench}   & 6,678   & 3,763 & 473 & 1.8 & \textcolor{red}{\xmark} & \textcolor{red}{\xmark} & \textcolor{orange!80!black}{Partial} \\
NExT-QA \citep{xiao2021next}          & 8,564   & 1,000 & 39.5 & 8.6 & \textcolor{red}{\xmark} & \textcolor{red}{\xmark} & \textcolor{red}{\xmark} \\
MSRVTT-QA  \citep{xu2016msr}       & 72,821  & 2,990 & 15.2 & 24.4  & \textcolor{red}{\xmark} & \textcolor{red}{\xmark} & \textcolor{red}{\xmark} \\
MSVD-QA \citep{xu2016msr}          & 13,157  & 504  & 9.8  & 26.1  & \textcolor{red}{\xmark} & \textcolor{red}{\xmark} & \textcolor{red}{\xmark} \\
STAR \citep{wu2star}           & 7,098   & 914  & 11.9 & 7.8 & \textcolor{red}{\xmark} & \textcolor{red}{\xmark} & \textcolor{red}{\xmark} \\
MVBench \citep{li2024mvbench}         & 4,000   & 3,641 & 16   & 1.1 & \textcolor{red}{\xmark} & \textcolor{red}{\xmark} & \textcolor{red}{\xmark} \\
TemporalBench \citep{cai2024temporalbench}    & 10,000  & 2000 & --   & 5 & \textcolor{red}{\xmark} & \textcolor{red}{\xmark} & \textcolor{orange!80!black}{Partial} \\
HV-MMBench \citep{cai2025hv} & 8,700 & 1,200 & -- & 7.25 & \textcolor{green!60!black}{\cmark} & \textcolor{red}{\xmark} & \textcolor{red}{\xmark} \\
\midrule
\textbf{FineBench (Ours)} & 199,420 & 64 & 900 & 3115.94 & \textcolor{green!60!black}{\cmark} & \textcolor{green!60!black}{\cmark} & \textcolor{green!60!black}{\cmark} \\
\bottomrule
\end{tabular}
\end{adjustbox}
\end{table*}
\section{Related Work}
\label{related}
Our work on \MethodName~builds upon extensive research in Video Question Answering (VQA) and the rapid advancements in Vision-Language Models (VLMs).

\noindent \textbf{Video Question Answering Datasets}. VQA evaluates video understanding via question answering. While numerous datasets exist, early influential ones like MSRVTT-QA \citep{xu2016msr} and ActivityNet-QA \citep{yu2019activitynet} often lacked dense spatial or temporal grounding, limiting fine-grained evaluation (Table~\ref{tab:vqa_comparison}). Subsequent datasets focused on deeper reasoning (e.g., NExT-QA \citep{xiao2021next}, STAR \citep{wu2star}) or specialized domains like egocentric video (EgoSchema \citep{mangalam2023egoschema}). Recent benchmarks (e.g., MovieChat \citep{song2024moviechat}, MVBench \citep{li2024mvbench}, TemporalBench \citep{cai2024temporalbench}, and MovieCORE \citep{faure2025moviecorecognitivereasoningmovies}) address various aspects like long videos or temporal reasoning. Some benchmarks explicitly emphasize human-centric evaluation. HumaniBench \citep{raza2025humanibench} focuses on human-centered AI principles such as fairness and empathy through image tasks, whereas HumanVBench \citep{zhou2024humanvbench} explores human-centric video understanding with synthetic data pipelines targeting emotion perception and speech–visual alignment. However, a gap remains for evaluating fine-grained human action understanding with dense grounding, particularly in complex scenes. \MethodName~addresses this gap by providing large-scale QA with dense \emph{spatial and temporal grounding of human actions and interactions} in long videos (avg. 900s), facilitating rigorous evaluation of precise human behavior localization and comprehension.


\noindent \textbf{Vision-Language Models (VLMs)}. Vision-Language Models (VLMs), integrating vision encoders and LLMs, have revolutionized cross-modal understanding with early works such as LlaVA \citep{liu2023visual}, MiniCPM-v2.6 \citep{yao2024minicpm}, and more recently, InternVL-2.5 \citep{chen2024expanding} and Qwen2.5-VL \citep{bai2025qwen2}. Extending this to video, recent VLMs like, mPlugOwl-3 \citep{ye2024mplug3}, and HERMES \citep{faure2024hermestemporalcoherentlongformunderstanding} handle temporal information to perform video tasks, including video captioning and VQA. Despite their capabilities, our analysis (Section~\ref{subsec:assess} and Section~\ref{subsec:why}) reveals significant challenges for these models in fine-grained video understanding, particularly concerning spatial localization in complex scenes and interpreting nuanced human actions and interactions. This underscores the need for human-centric benchmarks like \MethodName.

\section{\MethodName}
\label{sec:method}
To effectively evaluate Vision-Language Models' (VLMs) capacity for understanding nuanced visual content, we first delineate the characteristics of fine-grained video understanding as distinct from the general video understanding typically assessed by existing VQA datasets (Section \ref{subsec:overview}) and overview \MethodName. Section \ref{subsec:creation} then elaborates on our data creation and annotation process. Subsequently, we present extensive experiments benchmarking current VLMs to assess their proficiency in fine-grained video comprehension (Section \ref{subsec:assess}). Finally, Section \ref{subsec:why} examines the primary reasons these models struggle with such a task, providing insights for performance enhancements.

\begin{table*}[htbp]
    \centering
    \small
    \begin{minipage}[!b]{0.52\linewidth}
        \centering
        \caption{Key Statistics of \MethodName.}
        \label{tab:dataset_stats}
        \sisetup{group-separator={,}}
        \begin{tabular}{@{}lr@{}}
            \toprule
            Statistic & Value \\
            \midrule
            Total Questions           & \num{199420} \\
            Unique Videos             & \num{64} \\
            Avg. Annotated Frames/Video  & \num{785} \\
            \midrule
            \multicolumn{2}{l}{\textit{Category Distribution:}} \\
            \quad Person Movement     & \num{94330} (47.30\%) \\
            \quad Person Interaction  & \num{70140} (35.17\%) \\
            \quad Object Manipulation & \num{34950} (17.53\%) \\
            \midrule
            \multicolumn{2}{l}{\textit{Question Composition:}} \\
            \quad Single Actions   & \num{158625} (79.54\%) \\
            \quad Combined Actions & \num{40795} (20.46\%) \\
            \bottomrule
        \end{tabular}
    \end{minipage}
    \hfill
    \begin{minipage}[!b]{0.45\linewidth}
        \centering
        \includegraphics[width=\linewidth]{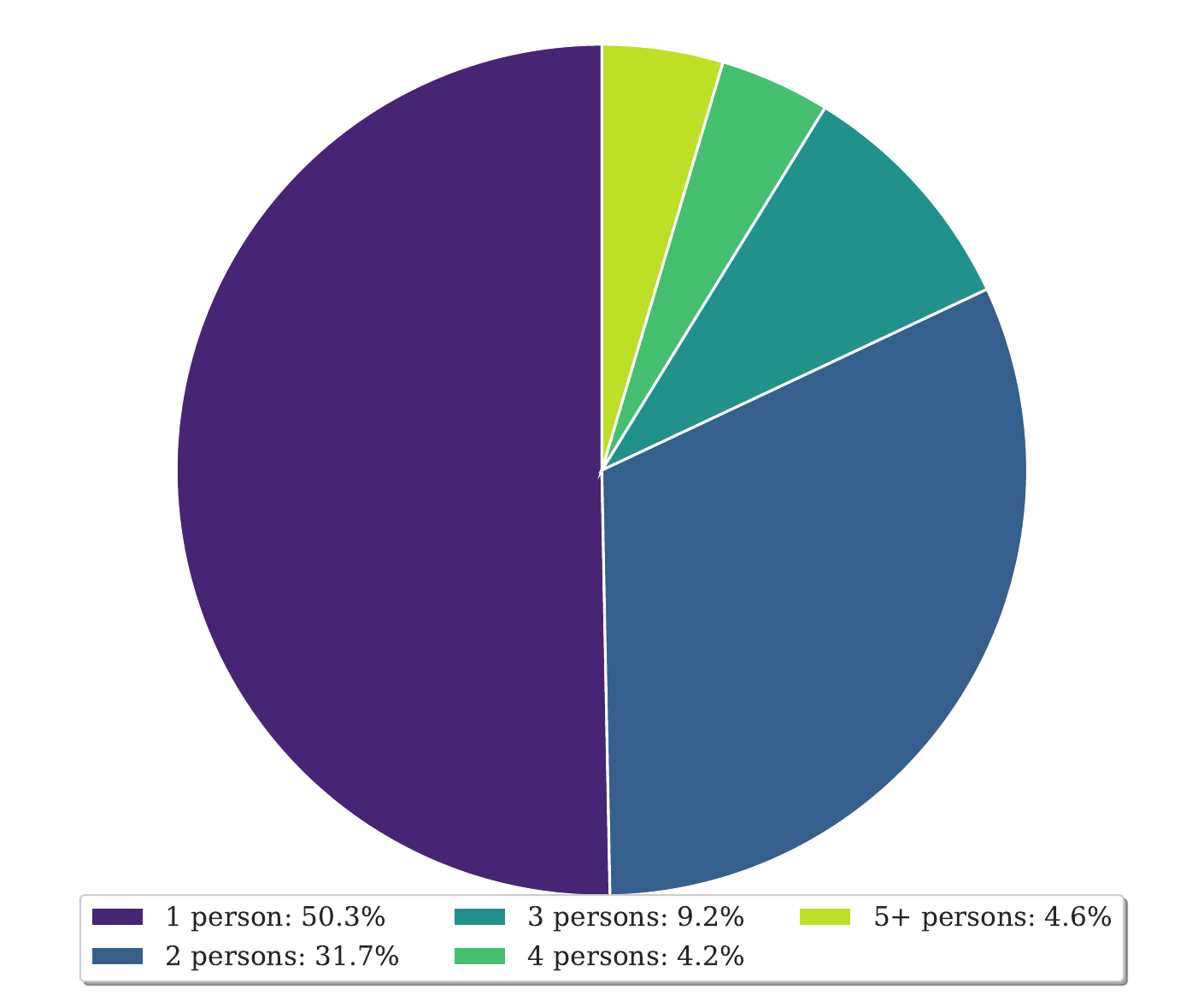}
        \captionof{figure}{Distribution of Annotated Persons per Keyframe.}
        \label{fig:person_dist}
    \end{minipage}
\end{table*}

\subsection{Overview of \MethodName}
\label{subsec:overview}
\textbf{Fine-grained video understanding} represents a crucial yet relatively underexplored facet of video-language models (VLMs). Unlike general video understanding tasks that focus on broad concepts, scene recognition, or high-level activities, fine-grained understanding requires models to perceive and reason about subtle visual details, momentary actions, and precise object interactions within video frames. A fine-grained \textit{human-centric} VQA dataset, in particular, must offer comprehensive coverage of all observable human behaviors. This includes not only the subject’s body pose and movement, but also their interactions with objects (\textit{person-object interactions}) and with other individuals (\textit{person-person interactions}). Figure~\ref{fig:benchmark_overview}(a) illustrates this diversity by showcasing QA examples across different reasoning types supported by \MethodName, from posture recognition to complex social interactions. Figure~\ref{fig:benchmark_overview}(b) highlights the temporal and spatial granularity required, where action labels evolve across frames and demand fine discrimination between visually similar behaviors.

\noindent \textbf{A ``Fine-grained" video understanding system} must posess the ability to distinguish between visually similar activities that share common attributes. In our context, this includes disambiguation (between actions such as ``carrying" vs. ``lifting" an object), temporal precision (identifying when actions start/end), spatial attention (focusing on the relevant regions of a frame), and contextual reasoning (understanding actions in relation to the environment).

\noindent \textbf{The importance of fine-grained understanding for VLMs} becomes evident when considering practical applications. In privacy-preserving ambient intelligence, general understanding might merely identify ``several people in a room," whereas fine-grained perception can distinguish whether individuals are ``standing in conversation," ``reaching for objects," or ``exhibiting signs of distress." For assisted living monitoring, fine-grained understanding allows systems to differentiate between ``a person deliberately sitting down" versus ``a person losing balance and falling"—a critical distinction for emergency response. Similar examples exist across human-robot interaction and everyday activities, positioning fine-grained video understanding as a fundamental capability that VLMs must possess to function effectively in complex real-world scenarios where subtle distinctions carry significant meaning.

\textbf{Our human-centric fine-grained video benchmark, \MethodName}, is structured as a multiple-choice video question answering (VQA) dataset, where each question is accompanied by four candidate answers, only one of which is correct. It contains a total of \textbf{199{,}420 QA pairs}, making it one of the largest VQA datasets. While the dataset relies on \textbf{64 unique videos} derived from the AVA dataset, these are highly dense 15-minute movie clips. Questions are densely linked to an average of \textbf{785 unique keyframes} per video, enabling detailed probing of model understanding at the second level. Unlike existing VQA datasets that focus on general comprehension or sparse annotation across many short clips, \MethodName~offers an average of \textbf{3{,}100 QA pairs per video}. This design choice explicitly prioritizes the depth and density of spatio-temporal grounding over superficial breadth, ensuring the benchmark thoroughly tests nuanced reasoning over extended video durations and fostering both local and holistic reasoning. 

Table~\ref{tab:dataset_stats} summarizes the key statistics of \MethodName. The dataset spans three broad conceptual domains—\textit{movement}, \textit{human interaction}, and \textit{object manipulation}—which guide the diversity of visual reasoning required. \textbf{Over 20\% of QA pairs involve combined activities}, testing compositional reasoning where multiple visual cues must be integrated. Figure~\ref{fig:person_dist} shows that nearly half the frames contain multiple annotated persons, emphasizing the fine-grained nature of the interactions present in \MethodName. These properties (along with those in Table~\ref{tab:vqa_comparison}) make \MethodName{} the first benchmark explicitly designed to test VLMs' fine-grained human-centric video understanding ability, where success requires precision in space, time, and context. 

\subsection{Dataset Creation Process}
\label{subsec:creation}

The construction of \MethodName~leverages the human-annotated action classes and bounding boxes provided by the AVA dataset \citep{gu2018ava}. Our methodology integrates three core components: (1) systematic question generation using predefined templates, (2) a principled distractor selection strategy, and (3) spatial reasoning for subject disambiguation and subject-specific QA generation.

\subsubsection{Question Template Design and Instantiation}
\label{subsubsec:templates}

We design a structured set of question templates categorized by the nature of the action being queried. Specifically, 23 templates were created for \textit{person movement} actions (e.g., ``How would you describe the movement of \textit{\{person\}}?''), 21 templates for \textit{object manipulation} actions (e.g., ``How is \textit{\{person\}} interacting with the object?''), and 25 templates for \textit{person interaction} actions (e.g., ``What social interaction is \textit{\{person\}} engaged in?''). To anchor these questions visually and ensure clarity, the placeholder \textit{\{person\}} within each template is instantiated using spatial descriptors derived dynamically from bounding box positions. Phrases such as ``the leftmost person'' or ``the person in the center'' are employed to unambiguously refer to the specific individual relevant to the question within the video frame.

Our reliance on template-based generation, as opposed to free-form LLM generation, is a deliberate design choice to prevent LLM hallucinations during dataset construction. By strictly binding questions and answers to rigorously annotated human labels from AVA, we ensure that \MethodName~strictly measures visual perception and spatial reasoning capabilities rather than VLMs' language priors.

\begin{table*}[htpb]
    \centering
    \caption{\textbf{Performance of 15 Vision-Language Models (VLMs) on \MethodName{}}. Proprietary models evaluated on a representative subset--comprising 7 representative videos and totaling 20,143 questions--are shown at the top. Open models are evaluated on both the subset and the full dataset. The best full-dataset open score is \textbf{bolded} and the second-best \underline{underlined}. [P.: Person; Obj.: Object]}
    \label{tab:vlm_performance}
    \sisetup{
        table-format=2.1,
        table-number-alignment = center
    }
    \begin{tabular}{@{}l r S S S S @{}} 
        \toprule
         & Size & {P. Movement} & {P. Interaction} & {Obj. Manipulation} & {Avg.} \\
        \midrule
        Random Choice             & --   & 25.0 & 25.0 & 25.0 & 25.0 \\
        \midrule
        \multicolumn{6}{@{}c}{\textit{Subset Evaluation}} \\
        \midrule
        GPT-4o {\scriptsize {(2024/08/26)}} \citep{gpt4o}            & --   & 70.9 & 73.9 & 84.4 & 74.3 \\
        GPT-5-mini {\scriptsize {(2025/08/07)}} \citep{gpt5} & -- & \textbf{75.9} & \textbf{75.3} & 85.3 & \textbf{77.4} \\
        Gemini-1.5-Flash \citep{team2024gemini} & --   & 71.2 & 66.8 & 81.9 & 71.6 \\
        Gemini-2.0-Flash \citep{team2024gemini} & --   & \textbf{75.9} & 68.7 & \textbf{86.3} & 75.2 \\
        SmolVLM \citep{marafioti2025smolvlm} & 2B   & 48.5 & 48.0 & 80.0 & 53.9 \\
        MiniCPM-2.6 \citep{yao2024minicpm} & 8B   & 49.5 & 57.4 & 84.8 & 58.4 \\
        mPlugOwl-3 \citep{ye2024mplug3}    & 7B   & 47.9 & 55.8 & 84.0 & 56.6 \\
        \midrule
        \multicolumn{6}{@{}c}{\textit{Full Dataset Evaluation}} \\
        \midrule
        InternVL-2.5 \citep{chen2024expanding} & 1B   & 33.8 & 40.2 & \textbf{79.6} & 44.1 \\
        SmolVLM \citep{marafioti2025smolvlm}    & 2B   & 47.9 & 50.5 & 71.0 & 52.9 \\
        Qwen2.5-VL \citep{bai2025qwen2}     & 3B   & 58.0 & 57.5 & 73.2 & 60.5 \\
        BLIP-3 \citep{xue2024xgen}         & 4B   & 34.3 & 58.6 & 64.9 & 48.2 \\
        InternVL-2.5 \citep{chen2024expanding} & 4B   & 61.4 & 58.6 & 78.1 & 63.3 \\
        mPlugOwl-2 \citep{ye2024mplug2}     & 7B   & 57.6 & 49.2 & \underline{78.5} & 58.3 \\
        mPlugOwl-3 \citep{ye2024mplug3}     & 7B   & 48.9 & 54.8 & 75.2 & 55.6 \\
        MiniCPM-2.6 \citep{yao2024minicpm}  & 8B   & 56.2 & 56.5 & 72.8 & 59.2 \\
        LLaVA-OV \citep{li2024llava}        & 7B   & 53.3 & 60.4 & 69.6 & 58.6 \\
        InternVL-2.5 \citep{chen2024expanding} & 8B   & \underline{66.8} & \underline{62.1} & 78.1 & \underline{67.1} \\
        Qwen2.5-VL \citep{bai2025qwen2}     & 7B   & \textbf{70.7} & \textbf{63.8} & 73.9 & \textbf{68.8} \\
        \bottomrule
    \end{tabular}
\end{table*}

\subsubsection{Distractor Generation Strategy}
\label{subsubsec:distractors}

For each annotated action instance in AVA v2.2, we generate a corresponding multiple-choice question. The process begins by classifying the ground truth action into one of the three categories: person movement, object manipulation, or person interaction. A question template is then randomly selected from the pool corresponding to that action category. Plausible distractors (incorrect answer options) are generated using a two-tiered approach. The primary strategy involves selecting actions that are semantically similar to the correct answer, based on a predefined similarity mapping. For example, actions like ``hand wave'', ``hand clap'', and ``hand shake'' are considered semantically close and may serve as distractors for one another, thereby increasing the question's difficulty. If no sufficiently similar actions are found via this mapping, a fallback strategy is employed: distractors are randomly selected from the same broad action category (e.g., other person movement actions) to maintain contextual relevance. In scenarios where an individual is annotated with multiple concurrent actions belonging to the same category, we formulate compound questions (e.g., reflecting simultaneous actions like ``listening to and watching a person'') and select appropriate distractors.

\subsubsection{Spatial Referencing and Disambiguation}
\label{subsubsec:spatial}
To enable precise questioning about specific individuals within a scene, especially when multiple people are present, we implement a dynamic spatial referencing system based on bounding box locations. When only one or two individuals are detected, relative positional terms (e.g., ``the person on the left'', ``the person on the right'') are used for disambiguation. For scenes containing three or more individuals, ordinal references (e.g., ``the second person from the left'') are generated to ensure clarity. This ensures that the generated questions unambiguously target the intended person.

\subsection{Do VLMs Exhibit Fine-Grained Video Understanding?}
\label{subsec:assess}

To evaluate whether current Vision-Language Models can perform fine-grained human-centric video understanding, we benchmark a diverse set of proprietary and open-source models using \MethodName{}, integrated into the VLMEvalkit \citep{duan2024vlmevalkit} library. Due to the high cost of querying proprietary APIs at scale, we provide results on two tiers: a representative subset (7 videos, 20,143 QAs) and the full dataset for open models only. The results are shown in Table~\ref{tab:vlm_performance}.

Proprietary models, notably GPT-5-mini \citep{gpt5} and Gemini-2.0-Flash \citep{team2024gemini}, demonstrate strong performance on the representative subset, substantially outperforming open models evaluated on the same data. This suggests these proprietary models possess stronger spatio-temporal reasoning and fine-grained human activity disambiguation capabilities, likely due to large-scale pretraining and robust multimodal pipelines. Crucially, as highlighted in Table~\ref{tab:vlm_performance}, the performance of open-source models on the subset closely matches their performance on the full dataset (e.g., MiniCPM scores 58.4\% on the subset vs. 59.2\% on the full dataset; similar tight tracking is observed for InternVL-2.5 and Qwen2.5-VL placeholders). This directly proves that the subset serves as a highly representative split, allowing for robust direct comparisons between closed-source APIs and open-source models without requiring the prohibitive costs of full-dataset evaluations for proprietary models. 

In contrast, open models exhibit wide variability and underwhelming accuracy on the full dataset. The top open model, Qwen2.5-VL (7B) \citep{bai2025qwen2}, achieves 68.8\%, but most models cluster around 55–60\%, and a few perform near chance level on Person Movement-related questions. These gaps indicate that current open VLMs struggle with fine-grained temporal cues, subtle interactions, and compositional reasoning—core challenges posed by \MethodName{}. Such results highlight a critical gap in the open ecosystem and a need for progress in training methods, architectures, and benchmarks tailored for fine-grained human-centric video comprehension.

\noindent\textit{\textbf{Finding 1:} Current VLMs exhibit a clear performance divide across action types. They handle object-centric tasks well but fall significantly short on human-centric reasoning, revealing that fine-grained human activity understanding remains an open challenge.}

\subsection{Why Do VLMs Struggle With Fine-Grained Video Understanding?}
\label{subsec:why}

\begin{figure*}[t!] 
    \centering 

    \begin{subfigure}[!t]{0.31\textwidth} 
        \centering
        \includegraphics[width=\linewidth]{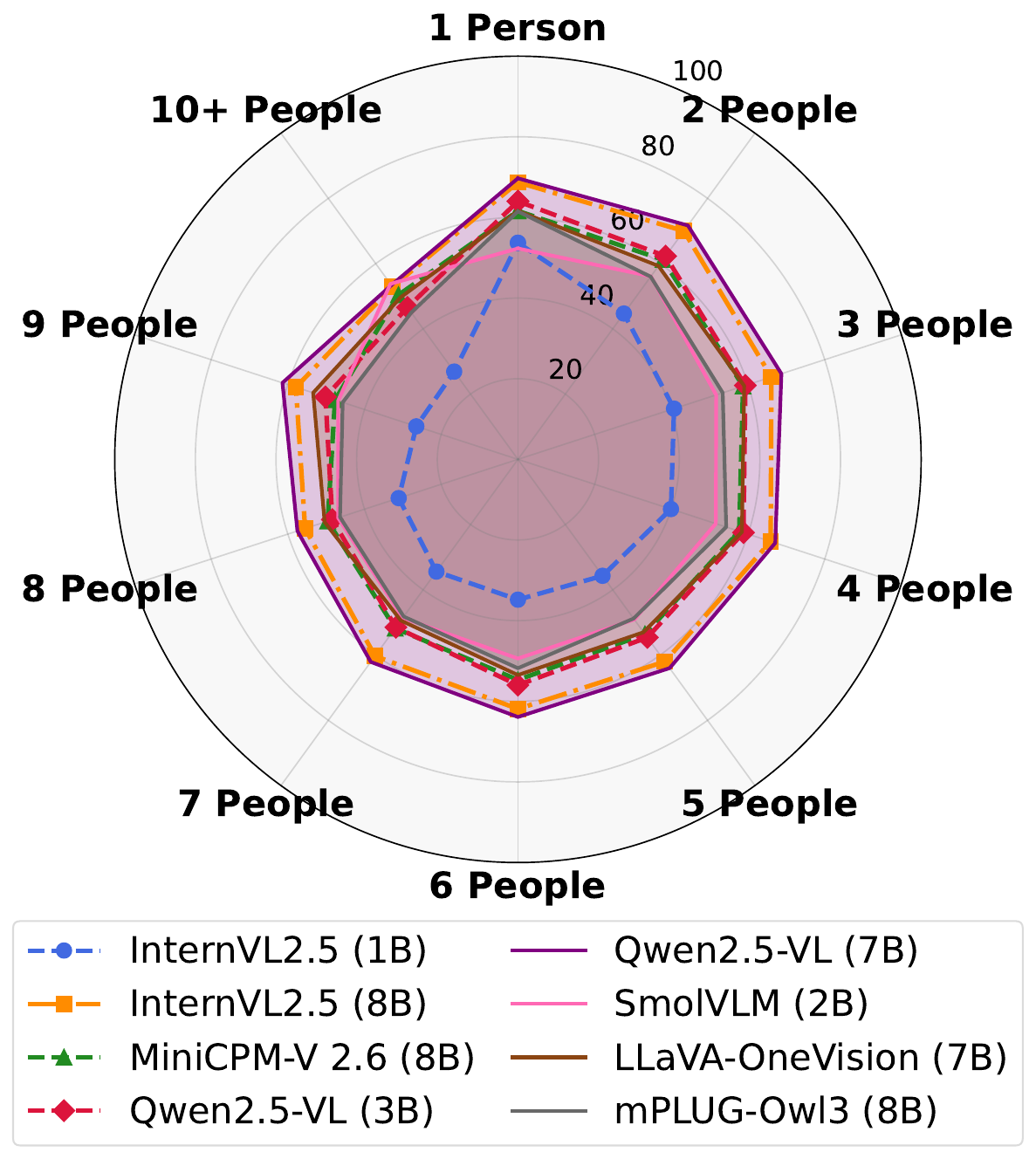}
        \caption{Acc. per number of person}
        \label{fig:radar_persons}
    \end{subfigure}
    \hfill 
    \begin{subfigure}[!t]{0.33\textwidth} 
        \centering
        \includegraphics[width=\linewidth]{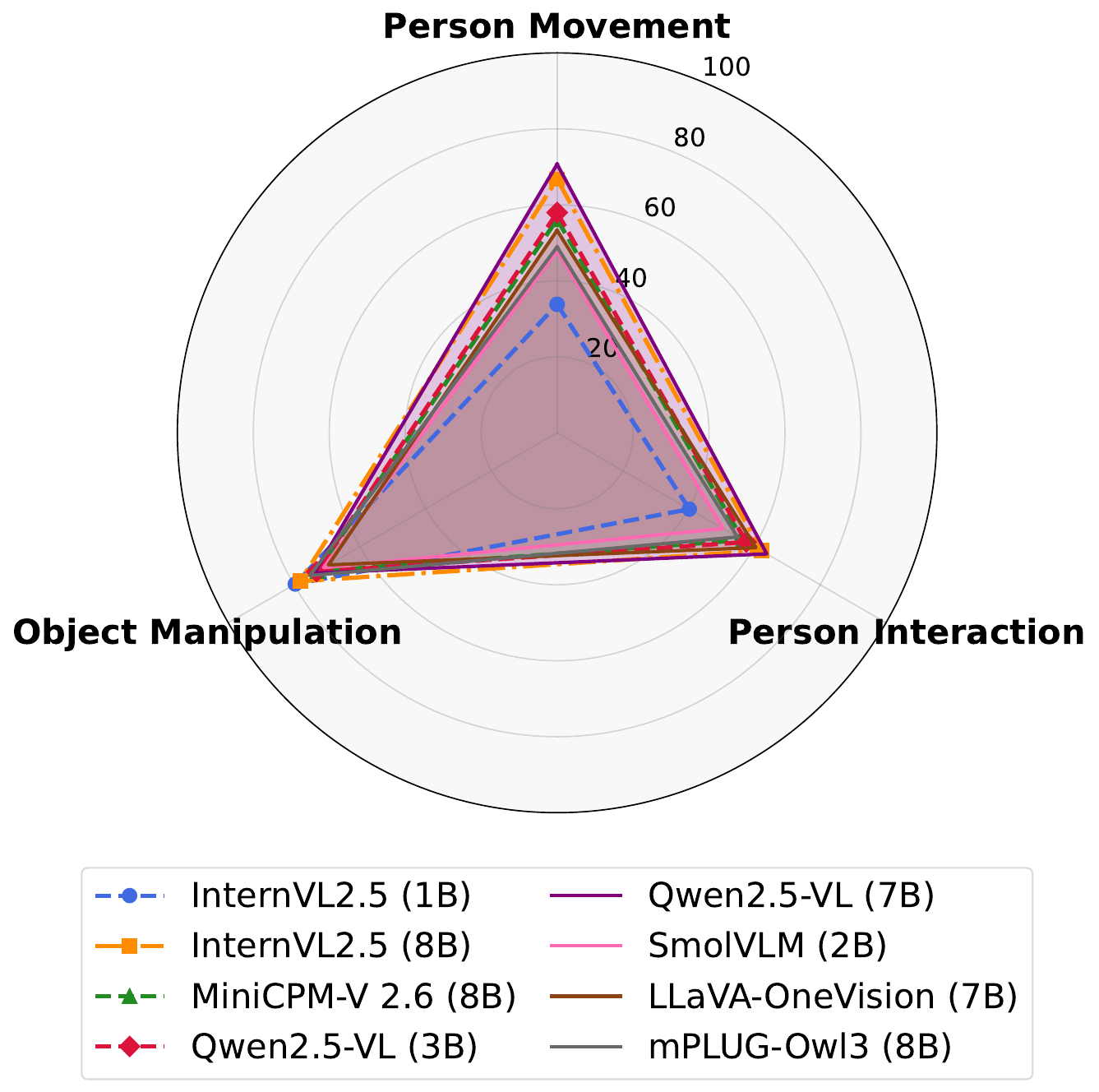}
        \caption{Acc. per question category}
        \label{fig:radar_categories} 
    \end{subfigure}
    \hfill 
    \begin{subfigure}[!b]{0.35\textwidth} 
        \centering
        \includegraphics[width=\linewidth]{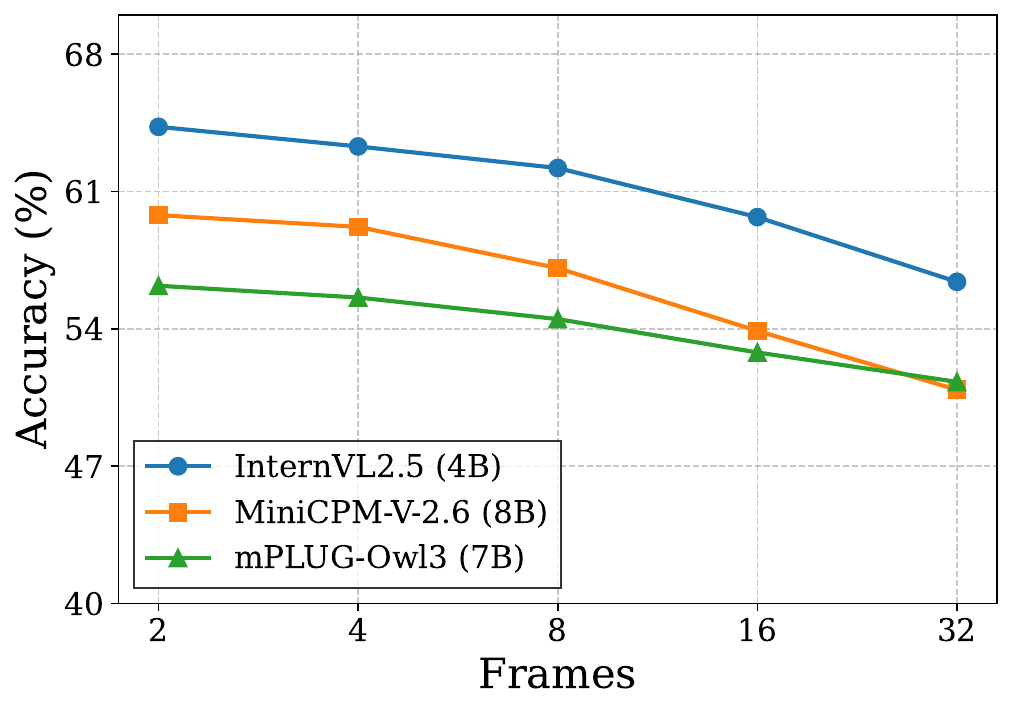}
        \caption{Accuracy per the number of frames}
        \label{fig:acc_vs_frame}
    \end{subfigure}

    \caption{VLM performance analysis on \MethodName~detailing accuracy variations. (a) Performance degradation with increasing number of persons in the scene. (b) Performance differences across action categories, with Person Movement being consistently lower. (c) Performance degradation with increasing number of frames.}
    \label{fig:radar_analysis} 
\end{figure*}

Having established that current Vision-Language Models (VLMs) underperform on fine-grained video understanding tasks (Section~\ref{subsec:assess}), we investigate the underlying reasons by dissecting their performance. Our analysis focuses on two key aspects: the impact of scene complexity (number of persons) and the variation in performance across different action categories, visualized through radar charts in Figure \ref{fig:radar_persons} and Figure~\ref{fig:radar_categories}, respectively. Additionally, we investigate the influence of input context length to ascertain if insufficient visual information is a bottleneck, as shown in Figure~\ref{fig:acc_vs_frame}.

First, analyzing the accuracy relative to the number of people present (Figure~\ref{fig:radar_persons}) reveals a significant and consistent challenge for all evaluated VLMs. There is a clear trend of performance degradation as the number of individuals in the frame increases. For example, Qwen2.5-VL (7B), the top-performing model overall, has a peak accuracy of 71.7\% in scenes with 2 persons, but this accuracy drops to 53.4\% when 10 or more people are present. This decline is even more pronounced for smaller models like InternVL-2.5 (1B), which drops from 53.7\% to 26.9\%. This consistent decrease suggests that VLMs struggle significantly with spatial reasoning, target disambiguation, and relationship understanding in complex, multi-person scenarios. Identifying and tracking the specific actions of designated individuals becomes substantially harder amidst visual clutter and potential occlusions.

Second, examining performance across action categories (Figure~\ref{fig:radar_categories}) highlights another area of weakness. Models consistently demonstrate higher proficiency in identifying \textit{Object Manipulation} actions compared to \textit{Person Movement} and \textit{Person Interaction}. Across all tested models, accuracy for Object Manipulation typically ranges from 71\% to nearly 80\%, whereas accuracies for the other two categories are often considerably lower. For instance, InternVL-2.5 (8B) achieves 78.1\% on Object Manipulation but only 66.8\% on Person Movement and 62.1\% on Person Interaction. This disparity suggests that VLMs find it easier to recognize actions centered around distinct object interactions, which may offer clearer visual cues. Conversely, they appear less capable of interpreting the nuances of human kinematics involved in diverse movements and the complex, often subtle, cues defining social interactions between individuals. These person-centric categories demand a deeper understanding of human pose, gestures, and context that current models do not fully capture. We also isolate the impact of the vision components with a blind evaluation showing that Qwen2.5VL (7B), MiniCPM-v2.6 (8B), and InternVL-2.5 (8B) score only 43.5, 29.9, and 33.0, respectively, when blind.

Our key takeaway is that current open-source VLMs struggle with fine-grained video understanding primarily due to two challenges. First, they exhibit deficiencies in robust spatial reasoning and subject disambiguation, particularly as scene complexity (number of actors) increases. This makes it difficult to correctly attribute actions to the right individuals. Second, they find it harder to interpret and distinguish nuanced human-centric actions, especially subtle body movements and complex social interactions, compared to more visually salient object-related actions. These person-centered tasks require models to pick up on fine-grained visual details and temporal patterns of human behavior, which current architectures and training paradigms are not yet adept at. Addressing these limitations is key for advancing fine-grained human-centric video understanding.

\noindent\textit{\textbf{Finding 2:} Scene complexity is a critical bottleneck: The more people present, the harder it becomes for VLMs to correctly attribute actions.}

\begin{figure}[htbp]
    \centering
    \includegraphics[width=\linewidth]{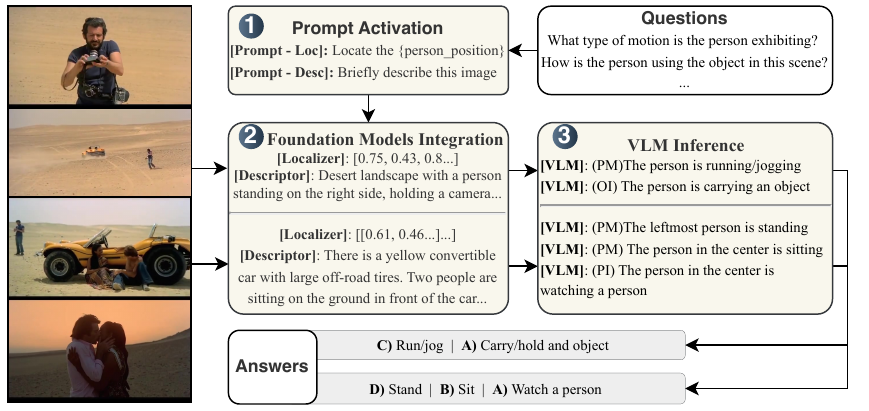}
    \caption{\textbf{Workflow of \AgentName}. It begins with (1) prompt activation for the Localizer and Descriptor. (2) The Localizer and Descriptor, both Foundation models, provide bounding box coordinates and textual captions. (3) Finally, the VLM uses this processed information during inference.}
    \label{fig:fineagent}
\end{figure}

\section{\AgentName}
\label{sec:fineagent}

Our error analysis in Section~\ref{subsec:why} identifies two primary obstacles hindering the fine-grained video understanding capabilities of current VLMs: (1) difficulties with spatial reasoning and subject disambiguation in multi-person scenes, and (2) a weaker grasp of nuanced human movements and interactions compared to object-centric actions. To address these limitations, we propose \textbf{\AgentName}, a modular framework designed to augment existing VLMs with spatial grounding and contextual information, thereby enhancing their fine-grained reasoning abilities.

\begin{table*}[htpb]
    \centering
    \caption{Performance gains with \AgentName~across different models.}
    \label{tab:fineagent}
    \sisetup{
        table-format=2.1,
        table-number-alignment = center
    }
    \begin{tabular}{@{}lcccc@{}}
        \toprule
        Model & P. Movement & P. Interaction & Obj. Manipulation & Avg. \\
        \midrule
        InternVL-2.5 (1B) \citep{chen2024expanding}  & 33.8 & 40.2 & 79.6 & 44.1 \\
        \quad + \textbf{\AgentName} & 
        \textbf{47.9} \scriptsize\textcolor{green!50!black}{(+14.1)} & 
        \textbf{44.2} \scriptsize\textcolor{green!50!black}{(+4.0)} & 
        \textbf{80.6} \scriptsize\textcolor{green!50!black}{(+1.0)} & 
        \textbf{52.4} \scriptsize\textcolor{green!50!black}{(+8.3)} \\
        \midrule
        Qwen2.5-VL (7B) \citep{bai2025qwen2} & 70.7 & 63.8 & 73.9 & 68.8 \\
        \quad + \textbf{\AgentName} & 
        \textbf{71.5} \scriptsize\textcolor{green!50!black}{(+0.8)} & 
        \textbf{64.1} \scriptsize\textcolor{green!50!black}{(+0.3)} & 
        \textbf{76.3} \scriptsize\textcolor{green!50!black}{(+2.4)} & 
        \textbf{69.7} \scriptsize\textcolor{green!50!black}{(+0.9)} \\
        \midrule
        mPlugOwl-3 (7B) \citep{ye2024mplug3} & 48.9 & 54.8 & 75.2 & 55.6 \\
        \quad + \textbf{\AgentName} & 
        \textbf{60.8} \scriptsize\textcolor{green!50!black}{(+11.9)} & 
        \textbf{57.8} \scriptsize\textcolor{green!50!black}{(+3.0)} & 
        \textbf{77.4} \scriptsize\textcolor{green!50!black}{(+2.2)} & 
        \textbf{62.7} \scriptsize\textcolor{green!50!black}{(+7.1)} \\
        \midrule
        MiniCPM-2.6 (8B) \citep{yao2024minicpm} & 56.2 & 56.5 & 72.8 & 59.2 \\
        \quad + \textbf{\AgentName} & 
        \textbf{60.6} \scriptsize\textcolor{green!50!black}{(+4.4)} & 
        \textbf{58.8} \scriptsize\textcolor{green!50!black}{(+2.3)} & 
        \textbf{76.3} \scriptsize\textcolor{green!50!black}{(+3.5)} & 
        \textbf{62.7} \scriptsize\textcolor{green!50!black}{(+3.5)} \\
        \bottomrule
    \end{tabular}
\end{table*}

\subsection{How does \textbf{\AgentName} Enhances Fine-grained Video Understanding?}
\label{subsec:how}

\textbf{\AgentName} enhances VLMs' fine-grained video understanding capabilities at inference time by integrating two complementary modules, designed to provide information that directly addresses the weaknesses identified in Section \ref{subsec:why}. The workflow of \AgentName{} is illustrated in Figure \ref{fig:fineagent}.

The first module is the \textbf{Localizer}, instantiated using EVFSam \citep{zhang2024evf}, a foundation model adept at visual grounding and referring segmentation. Given the video frames and the question, the Localizer provides the spatial location of the individual pertinent to the query. By supplying positional information, this module directly tackles the VLM's observed struggle with spatial reasoning and subject disambiguation in multi-person scenes. The Localizer thus assists the base VLM in anchoring its visual analysis to the correct subject, mitigating confusion in crowded environments.

The second module is the \textbf{Descriptor}. This component is responsible for generating captions for the relevant video frames. We utilize Qwen2.5-VL (7B) \citep{bai2025qwen2} as the Descriptor, due to its strong performance among open-source VLMs (Table~\ref{tab:vlm_performance}). The Descriptor addresses the VLM's weakness in interpreting subtle human-centric actions, particularly those categorized under Person Movement and Person Interaction (Figure~\ref{fig:radar_categories}). The generated captions provide semantic context and higher-level descriptions of potentially ambiguous activities. This augments the base VLM's understanding beyond raw visual features and aids in the interpretation of complex kinematics or social cues that might otherwise be missed. These two modules operate synergistically: the Localizer first identifies \textit{who} and \textit{where} the question is focused on, and then the Descriptor provides a textual interpretation of \textit{what} is happening. This structured, auxiliary information is then combined with the question and video input, and fed to the VLM to facilitate a more informed prediction.

The effectiveness of integrating \textbf{\AgentName} is demonstrated empirically in Table~\ref{tab:fineagent}. Augmenting various base VLMs—including InternVL-2.5 (1B) \citep{chen2024expanding}, Qwen2.5-VL (7B) \citep{bai2025qwen2}, mPLUG-Owl-3 (7B) \citep{ye2024mplug3}, and MiniCPM-2.6 (8B) \citep{yao2024minicpm} with \textbf{\AgentName} framework consistently yields performance improvements across all models and action categories on \MethodName{}. Notably, the improvements are often most pronounced in the challenging Person Movement and Person Interaction categories, directly addressing the identified weaknesses. For instance, augmenting the InternVL-2.5 (1B) model with \textbf{\AgentName} boosts its Person Movement accuracy by a substantial 14.1 percentage points and Person Interaction accuracy by 4.0 points, resulting in an overall 8.3-point increase in average accuracy. Similar positive trends, with varying magnitudes, are observed across the other models. This validates our hypothesis that by specifically targeting spatial grounding and providing richer contextual descriptions for human actions, \textbf{\AgentName} can successfully enhance the fine-grained video understanding capabilities of existing VLMs.

\noindent\textit{\textbf{Finding 3:} Explicitly providing spatial grounding and contextual descriptions at inference time consistently improves fine-grained video understanding, suggesting that targeted auxiliary information can compensate for architectural weaknesses without retraining.}

\begin{table} 
\small
    \centering
    \captionof{table}{Ablation study on \AgentName{} components. We report average accuracy (\%) on \MethodName{}. Each column corresponds to adding a specific module to the base VLM. Improvements over the base model are shown in \textcolor{green!50!black}{green}. $\dagger$ means InternVL2.5 (8B) is used as Descriptor.}
    \label{tab:ablation_study}
    \sisetup{table-format=2.1, table-number-alignment=center}
    \begin{tabular}{@{}l*{4}{S[table-format=2.1]@{\,}l}@{}}
        \toprule
        \multirow{2}{*}{Model} & 
        \multicolumn{2}{c}{+ Localizer} & 
        \multicolumn{2}{c}{+ Descriptor} & 
        \multicolumn{2}{c}{+ \AgentName} \\
        \cmidrule(lr){2-3} \cmidrule(lr){4-5} \cmidrule(l){6-7}
        & \multicolumn{1}{c}{Acc.} & \multicolumn{1}{c}{$\Delta$} 
        & \multicolumn{1}{c}{Acc.} & \multicolumn{1}{c}{$\Delta$}
        & \multicolumn{1}{c}{Acc.} & \multicolumn{1}{c}{$\Delta$} \\
        \midrule
        mPlugOwl-3 (7B) & 58.4 & \scriptsize\textcolor{green!50!black}{(+2.8)} 
                        & 62.5 & \scriptsize\textcolor{green!50!black}{(+6.9)} 
                        & \textbf{63.7} & \scriptsize\textcolor{green!50!black}{(+8.1)} \\
        Qwen2.5-VL (7B) & 69.3 & \scriptsize\textcolor{green!50!black}{(+0.5)} 
                        & 69.5 & \scriptsize\textcolor{green!50!black}{(+0.7)} 
                        & \textbf{69.7} & \scriptsize\textcolor{green!50!black}{(+0.9)} \\
        Qwen2.5-VL (7B) $\dagger$ & 69.3 & \scriptsize\textcolor{green!50!black}{(+0.5)} 
                        & \textbf{69.9} & \scriptsize\textcolor{green!50!black}{(+1.1)} 
                        & \textbf{70.2} & \scriptsize\textcolor{green!50!black}{(+1.4)} \\
        \bottomrule
    \end{tabular}
\end{table}

\subsection{Importance of \AgentName~Components}
\label{subsec:ablation}

Table~\ref{tab:ablation_study} ablates the contribution of each module. The Localizer alone yields modest but consistent gains (+2.8\% for mPlugOwl-3, +0.5\% for Qwen2.5-VL), confirming that explicit spatial grounding helps subject disambiguation in multi-person scenes. The Descriptor contributes more substantially for mPlugOwl-3 (+6.9\%), but minimally for Qwen2.5-VL (+0.7\%)---an expected result, since the Descriptor itself is powered by Qwen2.5-VL and thus offers little additional signal to the same backbone. Swapping in InternVL-2.5 (8B) as Descriptor ($\dagger$) recovers this gap (+1.1\%), further supporting this explanation. Combined, both modules act synergistically: the total gain exceeds the sum of individual contributions for both models.

\noindent\textit{\textbf{Finding 4:} Spatial grounding and semantic context are complementary: combining both yields synergistic gains, underscoring the importance of jointly addressing \textit{where} and \textit{what} in activity understanding.}

\section{Conclusion}
\label{sec:conclusion}
We introduce \MethodName{}, a densely annotated benchmark of \num{199420} QA pairs probing fine-grained, human-centric video understanding. Our evaluation exposes two systematic weaknesses in current open-source VLMs: poor spatial reasoning in multi-person scenes, and limited sensitivity to subtle human movements and interactions. \AgentName{} directly targets these bottlenecks via spatial grounding and contextual captioning, yielding consistent gains across diverse architectures without retraining. We hope \MethodName{} serves as a rigorous testbed to drive future progress in this underexplored yet practically critical domain.
{
    \small
    \bibliographystyle{ieeenat_fullname}
    \bibliography{main}

@String(AAAI = {AAAI})

@inproceedings{song2024moviechat,
  title={Moviechat: From dense token to sparse memory for long video understanding},
  author={Song, Enxin and Chai, Wenhao and Wang, Guanhong and Zhang, Yucheng and Zhou, Haoyang and Wu, Feiyang and Chi, Haozhe and Guo, Xun and Ye, Tian and Zhang, Yanting and others},
  booktitle={Proceedings of the IEEE/CVF Conference on Computer Vision and Pattern Recognition},
  pages={18221--18232},
  year={2024}
}

@inproceedings{li2024mvbench,
  title={Mvbench: A comprehensive multi-modal video understanding benchmark},
  author={Li, Kunchang and Wang, Yali and He, Yinan and Li, Yizhuo and Wang, Yi and Liu, Yi and Wang, Zun and Xu, Jilan and Chen, Guo and Luo, Ping and others},
  booktitle={Proceedings of the IEEE/CVF Conference on Computer Vision and Pattern Recognition},
  pages={22195--22206},
  year={2024}
}

@article{fu2024video,
  title={Video-MME: The First-Ever Comprehensive Evaluation Benchmark of Multi-modal LLMs in Video Analysis},
  author={Fu, Chaoyou and Dai, Yuhan and Luo, Yondong and Li, Lei and Ren, Shuhuai and Zhang, Renrui and Wang, Zihan and Zhou, Chenyu and Shen, Yunhang and Zhang, Mengdan and others},
  journal={arXiv preprint arXiv:2405.21075},
  year={2024}
}

@article{mangalam2023egoschema,
  title={Egoschema: A diagnostic benchmark for very long-form video language understanding},
  author={Mangalam, Karttikeya and Akshulakov, Raiymbek and Malik, Jitendra},
  journal={Advances in Neural Information Processing Systems},
  volume={36},
  pages={46212--46244},
  year={2023}
}

@article{yao2024minicpm,
  title={MiniCPM-V: A GPT-4V Level MLLM on Your Phone},
  author={Yao, Yuan and Yu, Tianyu and Zhang, Ao and Wang, Chongyi and Cui, Junbo and Zhu, Hongji and Cai, Tianchi and Li, Haoyu and Zhao, Weilin and He, Zhihui and others},
  journal={arXiv preprint arXiv:2408.01800},
  year={2024}
}

@misc{faure2024hermestemporalcoherentlongformunderstanding,
      title={HERMES: temporal-coHERent long-forM understanding with Episodes and Semantics}, 
      author={Gueter Josmy Faure and Jia-Fong Yeh and Min-Hung Chen and Hung-Ting Su and Winston H. Hsu and Shang-Hong Lai},
      year={2024},
      eprint={2408.17443},
      archivePrefix={arXiv},
      primaryClass={cs.CV},
      url={https://arxiv.org/abs/2408.17443}, 
}

@misc{gpt4o,
    author = {OpenAI},
    title = {Hello GPT-4o},
    howpublished = {\url{https://openai.com/index/hello-gpt-4o/}},
    year = {2024},
    note = {[Accessed 01-11-2024]},
}

@misc{gpt5,
    author = {OpenAI},
    title = {Introducing GPT-5},
    howpublished = {\url{https://openai.com/index/introducing-gpt-5/}},
    year = {2025},
    note = {[Accessed 31-08-2025]},
}

@article{wu2024longvideobench,
  title={Longvideobench: A benchmark for long-context interleaved video-language understanding},
  author={Wu, Haoning and Li, Dongxu and Chen, Bei and Li, Junnan},
  journal={Advances in Neural Information Processing Systems},
  volume={37},
  pages={28828--28857},
  year={2024}
}

@inproceedings{yu2019activitynet,
  title={Activitynet-qa: A dataset for understanding complex web videos via question answering},
  author={Yu, Zhou and Xu, Dejing and Yu, Jun and Yu, Ting and Zhao, Zhou and Zhuang, Yueting and Tao, Dacheng},
  booktitle={Proceedings of the AAAI Conference on Artificial Intelligence},
  volume={33},
  number={01},
  pages={9127--9134},
  year={2019}
}

@inproceedings{xiao2021next,
  title={Next-qa: Next phase of question-answering to explaining temporal actions},
  author={Xiao, Junbin and Shang, Xindi and Yao, Angela and Chua, Tat-Seng},
  booktitle={Proceedings of the IEEE/CVF conference on computer vision and pattern recognition},
  pages={9777--9786},
  year={2021}
}

@inproceedings{xu2016msr,
  title={Msr-vtt: A large video description dataset for bridging video and language},
  author={Xu, Jun and Mei, Tao and Yao, Ting and Rui, Yong},
  booktitle={Proceedings of the IEEE conference on computer vision and pattern recognition},
  pages={5288--5296},
  year={2016}
}

@inproceedings{wu2star,
  title={STAR: A Benchmark for Situated Reasoning in Real-World Videos},
  author={Wu, Bo and Yu, Shoubin and Chen, Zhenfang and Tenenbaum, Joshua B and Gan, Chuang},
  booktitle={Thirty-fifth Conference on Neural Information Processing Systems Datasets and Benchmarks Track (Round 2)},
  year={2024}
}

@article{cai2024temporalbench,
  title={Temporalbench: Benchmarking fine-grained temporal understanding for multimodal video models},
  author={Cai, Mu and Tan, Reuben and Zhang, Jianrui and Zou, Bocheng and Zhang, Kai and Yao, Feng and Zhu, Fangrui and Gu, Jing and Zhong, Yiwu and Shang, Yuzhang and others},
  journal={arXiv preprint arXiv:2410.10818},
  year={2024},
  pages={},
}

@inproceedings{gu2018ava,
  title={Ava: A video dataset of spatio-temporally localized atomic visual actions},
  author={Gu, Chunhui and Sun, Chen and Ross, David A and Vondrick, Carl and Pantofaru, Caroline and Li, Yeqing and Vijayanarasimhan, Sudheendra and Toderici, George and Ricco, Susanna and Sukthankar, Rahul and others},
  booktitle={Proceedings of the IEEE conference on computer vision and pattern recognition},
  pages={6047--6056},
  year={2018}
}

@inproceedings{ye2024mplug2,
  title={mplug-owl2: Revolutionizing multi-modal large language model with modality collaboration},
  author={Ye, Qinghao and Xu, Haiyang and Ye, Jiabo and Yan, Ming and Hu, Anwen and Liu, Haowei and Qian, Qi and Zhang, Ji and Huang, Fei},
  booktitle={Proceedings of the ieee/cvf conference on computer vision and pattern recognition},
  pages={13040--13051},
  year={2024}
}

@article{ye2024mplug3,
  title={mplug-owl3: Towards long image-sequence understanding in multi-modal large language models},
  author={Ye, Jiabo and Xu, Haiyang and Liu, Haowei and Hu, Anwen and Yan, Ming and Qian, Qi and Zhang, Ji and Huang, Fei and Zhou, Jingren},
  journal={arXiv preprint arXiv:2408.04840},
  year={2024}
}

@article{chen2024expanding,
  title={Expanding performance boundaries of open-source multimodal models with model, data, and test-time scaling},
  author={Chen, Zhe and Wang, Weiyun and Cao, Yue and Liu, Yangzhou and Gao, Zhangwei and Cui, Erfei and Zhu, Jinguo and Ye, Shenglong and Tian, Hao and Liu, Zhaoyang and others},
  journal={arXiv preprint arXiv:2412.05271},
  year={2024}
}

@article{marafioti2025smolvlm,
  title={SmolVLM: Redefining small and efficient multimodal models},
  author={Marafioti, Andr{\'e}s and Zohar, Orr and Farr{\'e}, Miquel and Noyan, Merve and Bakouch, Elie and Cuenca, Pedro and Zakka, Cyril and Allal, Loubna Ben and Lozhkov, Anton and Tazi, Nouamane and others},
  journal={arXiv preprint arXiv:2504.05299},
  year={2025}
}

@article{xue2024xgen,
  title={xgen-mm (blip-3): A family of open large multimodal models},
  author={Xue, Le and Shu, Manli and Awadalla, Anas and Wang, Jun and Yan, An and Purushwalkam, Senthil and Zhou, Honglu and Prabhu, Viraj and Dai, Yutong and Ryoo, Michael S and others},
  journal={arXiv preprint arXiv:2408.08872},
  year={2024}
}

@article{li2024llava,
  title={Llava-onevision: Easy visual task transfer},
  author={Li, Bo and Zhang, Yuanhan and Guo, Dong and Zhang, Renrui and Li, Feng and Zhang, Hao and Zhang, Kaichen and Zhang, Peiyuan and Li, Yanwei and Liu, Ziwei and others},
  journal={arXiv preprint arXiv:2408.03326},
  year={2024}
}

@article{bai2025qwen2,
  title={Qwen2. 5-vl technical report},
  author={Bai, Shuai and Chen, Keqin and Liu, Xuejing and Wang, Jialin and Ge, Wenbin and Song, Sibo and Dang, Kai and Wang, Peng and Wang, Shijie and Tang, Jun and others},
  journal={arXiv preprint arXiv:2502.13923},
  year={2025}
}

@article{team2024gemini,
  title={Gemini 1.5: Unlocking multimodal understanding across millions of tokens of context},
  author={Team, Gemini and Georgiev, Petko and Lei, Ving Ian and Burnell, Ryan and Bai, Libin and Gulati, Anmol and Tanzer, Garrett and Vincent, Damien and Pan, Zhufeng and Wang, Shibo and others},
  journal={arXiv preprint arXiv:2403.05530},
  year={2024}
}

@article{liu2023visual,
  title={Visual instruction tuning},
  author={Liu, Haotian and Li, Chunyuan and Wu, Qingyang and Lee, Yong Jae},
  journal={Advances in neural information processing systems},
  volume={36},
  pages={34892--34916},
  year={2023}
}

@misc{faure2025moviecorecognitivereasoningmovies,
      title={MovieCORE: COgnitive REasoning in Movies}, 
      author={Gueter Josmy Faure and Min-Hung Chen and Jia-Fong Yeh and Ying Cheng and Hung-Ting Su and Yung-Hao Tang and Shang-Hong Lai and Winston H. Hsu},
      year={2025},
      eprint={2508.19026},
      archivePrefix={arXiv},
      primaryClass={cs.CL},
      url={https://arxiv.org/abs/2508.19026}, 
}

@inproceedings{duan2024vlmevalkit,
  title={Vlmevalkit: An open-source toolkit for evaluating large multi-modality models},
  author={Duan, Haodong and Yang, Junming and Qiao, Yuxuan and Fang, Xinyu and Chen, Lin and Liu, Yuan and Dong, Xiaoyi and Zang, Yuhang and Zhang, Pan and Wang, Jiaqi and others},
  booktitle={Proceedings of the 32nd ACM international conference on multimedia},
  pages={11198--11201},
  year={2024}
}

@misc{zhang2024evf,
  title={Evf-sam: Early vision-language fusion for text-prompted segment anything model},
  author={Zhang, Yuxuan and Cheng, Tianheng and Zhu, Lianghui and Hu, Rui and Liu, Lei and Liu, Heng and Ran, Longjin and Chen, Xiaoxin and Liu, Wenyu and Wang, Xinggang},
  journal={arXiv preprint arXiv:2406.20076},
  year={2024}
}

@article{raza2025humanibench,
  title={Humanibench: A human-centric framework for large multimodal models evaluation},
  author={Raza, Shaina and Narayanan, Aravind and Khazaie, Vahid Reza and Vayani, Ashmal and Chettiar, Mukund S and Singh, Amandeep and Shah, Mubarak and Pandya, Deval},
  journal={arXiv preprint arXiv:2505.11454},
  year={2025}
}

@article{zhou2024humanvbench,
  title={Humanvbench: Exploring human-centric video understanding capabilities of mllms with synthetic benchmark data},
  author={Zhou, Ting and Chen, Daoyuan and Jiao, Qirui and Ding, Bolin and Li, Yaliang and Shen, Ying},
  journal={arXiv preprint arXiv:2412.17574},
  year={2024}
}

@article{cai2025hv,
  title={HV-MMBench: Benchmarking MLLMs for Human-Centric Video Understanding},
  author={Cai, Yuxuan and Zhang, Jiangning and Gan, Zhenye and He, Qingdong and Hu, Xiaobin and Zhu, Junwei and Wang, Yabiao and Wang, Chengjie and Xue, Zhucun and He, Xinwei and others},
  journal={arXiv preprint arXiv:2507.04909},
  year={2025}
}
}


\end{document}